\documentclass[letterpaper, 10 pt, conference]{ieeeconf}  

\IEEEoverridecommandlockouts                              

\usepackage{amsmath}
\usepackage{graphicx}
\usepackage{capt-of}
\usepackage{amsfonts}
\usepackage{cleveref}
\usepackage{bbm, dsfont}
\usepackage{booktabs}
\usepackage{multicol}
\usepackage{multirow}
\usepackage{comment}
\usepackage{array}
\usepackage{xcolor}
\usepackage{xspace}
\usepackage{tikz}
\usepackage{subfig} 
\usepackage{xcolor} 
\usepackage{subcaption}

\usepackage{caption}
\makeatletter
\DeclareRobustCommand\onedot{\futurelet\@let@token\@onedot}
\def\@onedot{\ifx\@let@token.\else.\null\fi\xspace}

\makeatother

\title{\LARGE \bf
Meteorological data and Sky Images meets Neural Models for Photovoltaic Power Forecasting}

\author{Inés Montoya-Espinagosa$^{1}$ and Antonio Agudo$^{2}$, {\em Member, IEEE}
\thanks{This work has been supported by the project GreenVAR of the Fundación Ramón Areces.}
\thanks{$^{1}$Inés Montoya-Espinagosa is with the Universitat Pompeu Fabra, Barcelona, Spain.} \thanks{$^{2}$Antonio Agudo is with the Institut de Rob\`otica i Inform\`atica Industrial, CSIC-UPC, Barcelona, Spain.}
}%

\begin{document}

\maketitle

\vspace{-4mm}
\begin{abstract}
Due to the rise in the use of renewable energies as an alternative to traditional ones, and especially solar energy, there is increasing interest in studying how to address photovoltaic forecasting in the face of the challenge of variability in photovoltaic energy production, using different methodologies. This work develops a hybrid approach for short and long-term forecasting based on two studies with the same purpose. A multimodal approach that combines images of the sky and photovoltaic energy history with meteorological data is proposed. The main goal is to improve the accuracy of ramp event prediction, increase the robustness of forecasts in cloudy conditions, and extend capabilities beyond nowcasting, to support more efficient operation of the power grid and better management of solar variability. Deep neural models are used for both nowcasting and forecasting solutions, incorporating individual and multiple meteorological variables, as well as an analytical solar position. The results demonstrate that the inclusion of meteorological data, particularly the surface long-wave, radiation downwards, and the combination of wind and solar position, significantly improves current predictions in both nowcasting and forecasting tasks, especially on cloudy days. This study highlights the importance of integrating diverse data sources to improve the reliability and interpretability of solar energy prediction models.
\end{abstract}

\section{Introduction}

Photovoltaic (PV) solar energy has become a crucial element in the global transition to renewable energy~\cite{perez_gonzalo_2023}, and is projected to become the largest renewable source by 2030~\cite{ieaGlobalEnergy,perez_gonzalo_2024,dsa-ilora}. However, the variability of solar energy, mainly due to the dynamic nature of clouds, causes rapid fluctuations in irradiation resulting in an unstable energy production~\cite{Barnes2014,cozzi2020world}, which negatively impacts grid stability~\cite{10.1007/978-981-96-2468-3_32, Kim2024,Othman_2025}. This unpredictability makes intra-hour solar forecasting and short-term photovoltaic forecasting particularly difficult, yet essential for maintaining grid stability and ensure safe system operation~\cite{Lorenz2014, Lin2023, Zahraoui2024, Zhang2023, sun2019forecast, Mokri_2025}. To address these challenges, different studies aim to improve the accuracy of solar forecasts using sky image-based prediction models~\cite{feng2019opensolar, nie2023skippd}, some physical information such as solar irradiance and temperature~\cite{das2017svr} or perceptible water~\cite{ahmed2020review} were used. 

An approach to improve forecast accuracy involves using sky image-based prediction models, which leverage visual information to anticipate PV fluctuations more accurately than purely numerical and statistical models~\cite{Fabel2024, Zhang2023}. For instance, \cite{Zang2024} developed a two-stream network with PV energy-guided attention, achieving high accuracy but struggling with limited data and a lack of interpretability. Similarly, \cite{Fabel2024} used a deep-learning model for irradiance nowcasting and integrated it into a hybrid system with physics-based models and smart persistence, improving persistence models, but still facing difficulties predicting ramp events (sudden changes in power). Other studies have explored hybrid deep-learning architectures. For example, \cite{Kong2020} developed models using static and dynamic sky images, finding that their hybrid model achieved higher performance but generated blurry cloud boundary predictions. \cite{Kuo2022} used cloud coverage rates for higher accuracy but the method fails in rainy conditions and encounters limitations in RGB-based coverage calculations. To address fog-affected scenarios, \cite{Lu2023} introduced the aerosol scattering coefficient as an auxiliary feature, achieving notable improvements when combined with other algorithms based on support vector machines, long short-term memory (LSTM) models, and XGBoost. \cite{Hu2022} used cloud motion vectors for high accuracy but resulted in an overfitted model heavily relies on accurate cloud tracking. Collectively, these studies underscore the need for robust, multi-modal approaches that can adapt to diverse weather conditions. Meanwhile, \cite{Zhen2020} proposed a hybrid mapping model combining Convolutional Neural Networks (CNN) and LSTM, which outperformed standalone models in various metrics but required further improvement in generalizing to varied cloud dynamics. \cite{nie2023skippd} used CNNs with the Sky Images and Photovoltaic Power Dataset (SKIPP'D) for PV prediction and in~\cite{Berresheim_Agudo_2024} incorporated image-based estimations such as cloud and sun segmentation or sun position to increase the robustness of the method. However, these approaches faced key challenges, including the inability to learn all important patterns, reduced reliability during cloudy weather with rapid power changes, as well as issues with image brightness and cloud identification. The increasing use of solar energy into power grid has revealed a critical problem: solar energy production varies a lot due to the cloud’s motion and changing weather. While artificial-intelligence methods have helped with short-term solar predictions, current systems still have two key limitations: (1) efficiently predict fluctuations in solar irradiance (called ramp events), and (2) perform accurate forecasts for very short timeframe. These challenges have a direct impact on the grid stability and operational efficiency, since they need to perfectly match electricity supply with demand at all times.

To address the previous gaps, this work proposes a multi-modal hybrid methodology that combines ground-based sky images from SKIPP'D~\cite{nie2023skippd} with physical meteorological data from the ERA5 reanalysis~\cite{ecmwfECMWFReanalysis}. The objective is to enhance the prediction of ramp events, improve forecast robustness, extend capabilities beyond nowcasting, and provide a physical-aware interpretability to support more efficient and reliable power grid operation.

\section{Meteorological and visual information}

We now introduce the information we propose to use in the prediction of PV power. To this end, we consider the dataset SKIPP'D~\cite{nie2023skippd} for visual information and ERA5~\cite{ERA5} for meteorological one. 


\subsection{Visual information} 

The SKIPP'D dataset~\cite{nie2023skippd} comprises ground-based sky images captured at a frequency of 20 Hz and a high spatial resolution of 2,048$\times$2,048 pixels using a 360-degree fisheye camera. The recordings span the period from March 2017 to December 2019 and cover the time interval between 06:00 AM and 08:00 PM each day. In addition to the imagery, the dataset includes PV power generation data recorded at 1-minute intervals. Following the work described in~\cite{nie2023skippd,Berresheim_Agudo_2024}, we utilize 363,375 sky image samples with a temporal resolution of one minute and a downsampled spatial resolution of 64$\times$64 pixels (some instances are displayed in Fig.~\ref{fig:sunpos}), along with their corresponding PV output observations, and employing a 96\%/4\% split for training and testing, respectively. The dataset contains 20 selected days: 10 sunny and 10 cloudy. 

From that data, we derived an additional dataset for forecasting purposes, referred to as the Forecast dataset. This contains 137,489 samples, approximately half the size of the original, by applying a two-minute sampling interval each day. In this particular case, it has been employed a 94.25\%/5.75\% split for training and testing, respectively. It is worth noting that in this case sequences of 16 consecutive images (covering the past 15 minutes) and the corresponding 16 PV power values are used.

\subsection{Meteorological information} 

The ERA5~\cite{ERA5} reanalysis dataset from ECMWF~\cite{ecmwfECMWFReanalysis} provides a globally consistent, hourly record of meteorological variables from 1940 to present. Data are provided on a 0.25-degree grid (approximately 31 km in resolution). Next, we introduce some variables that will be later analyzed in PV power prediction: 

\subsubsection{Data Coverage and Extraction}
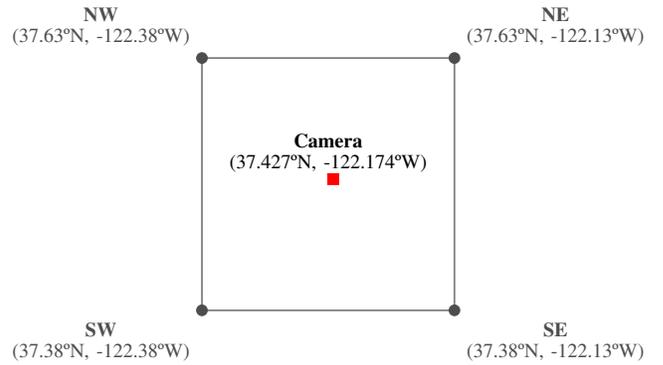
\begin{figure}[t!]
\centering
\resizebox{8.7cm}{!} {
\begin{tikzpicture}[scale=2, every node/.style={font=\small}]
  \coordinate (NW) at (-1,1);
  \coordinate (NE) at (1,1);
  \coordinate (SW) at (-1,-1);
  \coordinate (SE) at (1,-1);
  \coordinate (CAM) at (0,0);
  \draw[gray, thick] (NW) -- (NE) -- (SE) -- (SW) -- cycle;
    \filldraw[black!70] (NW) circle (1.2pt)
      node[above left=2pt] {\shortstack{\textbf{NW} \\ (37.63ºN, -122.38ºW)}};
    \filldraw[black!70] (NE) circle (1.2pt)
      node[above right=2pt] {\shortstack{\textbf{NE} \\ (37.63ºN, -122.13ºW)}};
    \filldraw[black!70] (SW) circle (1.2pt)
      node[below left=2pt] {\shortstack{\textbf{SW} \\ (37.38ºN, -122.38ºW)}};
    \filldraw[black!70] (SE) circle (1.2pt)
      node[below right=2pt] {\shortstack{\textbf{SE} \\ (37.38ºN, -122.13ºW)}};
  \draw[red, thick, fill=red] (CAM) rectangle ++(0.08,0.08);
    \node[above=2pt] at (CAM) {\shortstack{\textbf{Camera} \\ (37.427ºN, -122.174ºW)}};
\end{tikzpicture}}
\caption{\textbf{Latitude and longitude of the four ERA5~\cite{ERA5} grid points} surrounding the camera location in Stanford University. These points, spaced at 0.25° resolution (NW, NE, SW, SE), define the area from which ERA5~\cite{ERA5} variables are extracted. The red square indicates the real camera location for image acquisition.}
\label{fig:era5_grid}
\end{figure}

Meteorological data are extracted from four grid points (NW, NE, SW, SE) surrounding the camera location at Stanford University (37.427°N, -122.174°W), as shown in Fig.~\ref{fig:era5_grid}. The spatial coverage of this area is displayed in the figure.

\subsubsection{Meteorological variables}
Ten key variables have been extracted, categorized as (a summary of meteorological data is displayed in Table~\ref{tab:variable_characteristics}):
\begin{itemize}
    \item \textbf{Instantaneous analysis variables}: Total cloud cover (\texttt{tcc}), surface pressure (\texttt{sp}), and wind components (\texttt{Wind100} which is composed by \texttt{100u} and \texttt{100v}).
    \item \textbf{Forecast non-accumulative variable}: Wind gust (\texttt{i10fg}).
    \item \textbf{Forecast accumulative variables}: Radiation fluxes: (\texttt{ssrd}, \texttt{strd}, \texttt{str}, \texttt{tsr}, \texttt{fdir}).
\end{itemize}

\subsubsection{Meteorological data preprocessing}
\label{variables_meteo}

Three different preprocessing techniques have been applied based on variable type:
\begin{itemize}
    \item \textbf{Instantaneous analysis variables}: Converted from UTC to local time, extracted from the four grid points (NW, NE, SW, SE), aligned to image timestamps using forward/backward fill, and normalized. 
    \item \textbf{Forecast non-accumulative variable}: Real timestamps reconstructed using \texttt{time + step}, then processed similarly to analysis variables.
    \item \textbf{Forecast accumulative variables}: Converted from cumulative to hourly rates, then flattened, assigned real timestamps, and interpolated to 1-minute resolution.
\end{itemize}

Mean and standard deviation are computed on the training and validation split, forcing standard deviation to 1.0 to avoid division by zero. This normalization ensures that all variables contribute equally during model training, regardless of their original units and/or magnitudes.

\subsection{Sun Position}

Finally, we also propose the sun position as an additional source of information to improve model prediction by helping identify lighting patterns and atmospheric conditions linked to the solar cycle.

To this end, we propose an analytical way to infer the solar location by exploiting two fundamental solar angles:
\begin{itemize}
    \item Solar azimuth ($\alpha$): horizontal angle from true north ($0^\circ$–$360^\circ$).
    \item Solar elevation ($\epsilon$): angle between the horizon and the sun ($0^\circ$–$90^\circ$).
\end{itemize}

\begin{table}[t!]
    \centering
        \caption{\textbf{ERA5~\cite{ERA5} variables (\texttt{fc} = forecast; \texttt{an} = analysis) explanation.} The table shows, from left to right, a summary of the official parameter name of the meteorological variables, the full name, the source (if it is a analysis or forecast variable), whether the data has or not a step, the units, and the variable type (whether it is an accumulative variable in time or not).}
    \resizebox{8.6cm}{!} { 
    \begin{tabular}{c|l|c|c|c|l}
        \hline
        \textbf{Variable} & \textbf{Full name} & \textbf{Source} & \textbf{Step} & \textbf{Units} & \textbf{Variable type} \\
        \midrule
        \texttt{tcc} & Total Cloud Cover & an & No & Fraction (0–1) & Instantaneous \\
        \texttt{i10fg} & 10m Wind Gust & fc & Yes & m/s & Instantaneous \\
        \texttt{100u} & U component of wind at 100m & an & No & m/s & Instantaneous \\
        \texttt{100v} & V component of wind at 100m & an & No & m/s & Instantaneous \\
        \texttt{sp} & Surface Pressure & an & No & Pa & Instantaneous \\
        \texttt{strd} & Surface Thermal Radiation Downwards & fc & Yes & J/m² & Accumulative \\
        \texttt{ssrd} & Surface Solar Radiation Downwards & fc & Yes & J/m² & Accumulative \\
        \texttt{str} & Surface Net Thermal Radiation & fc & Yes & J/m² & Accumulative \\
        \texttt{tsr} & Top Net Solar Radiation & fc & Yes & J/m² & Accumulative \\
        \texttt{fdir} & Surface Solar Direct Radiation & fc & Yes & J/m² & Accumulative \\
        \bottomrule
    \end{tabular}}
    \label{tab:variable_characteristics}
\end{table}

These parameters were obtained via \texttt{pvlib}~\cite{F_Holmgren2018-mp}, a Python package specifically designed for solar energy system modeling. For the fisheye camera at Stanford University (latitude 37.427$^\circ$, longitude -122.174$^\circ$, oriented 194$^\circ$ from true north), the solar azimuth is first converted to a relative azimuth with respect to the camera's optical axis as:
\begin{equation}
\alpha_{\text{rel}} = (\alpha_{\text{sun}} - \alpha_{\text{cam}}) \,\, \textrm{with} \mod 360^\circ ,
\end{equation}
where $\alpha_{\text{rel}}$ is the corrected relative azimuth, $\alpha_{\text{sun}}$ is the solar azimuth and $\alpha_{\text{cam}} = 194^\circ$ indicates the camera's orientation. The relative azimuth is then converted to radians as $\theta = \textrm{deg2rad}(\alpha_{\text{rel}})$. 

The radial distance from the image center is calculated using a normalized zenith angle:
\begin{equation}
\theta_{\text{norm}} = \frac{90^\circ - \epsilon}{90^\circ},
\end{equation}
that it is used to infer the radial distance $r$ as:
\begin{equation}
r = R \cdot (\theta_{\text{norm}})^\gamma ,
\end{equation}
where $R$ is the maximum image radius (32 pixels for 64$\times$64 images) and $\gamma = 1.2$ denotes a non-linear correction factor.

\begin{figure}[t!]
    \centering
    \resizebox{8.6cm}{!} { 
    \begin{tabular}{@{}cccc@{}}
{\includegraphics[width=0.27\linewidth]{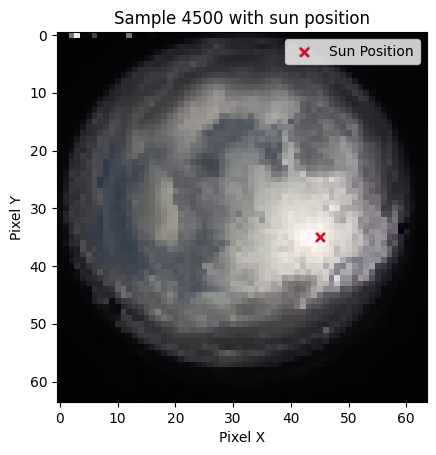}\label{fig:sub1}}&
{\includegraphics[width=0.27\linewidth]{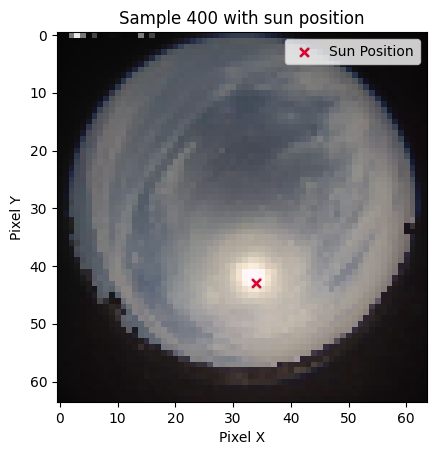}\label{fig:sub2}}&
{\includegraphics[width=0.27\linewidth]{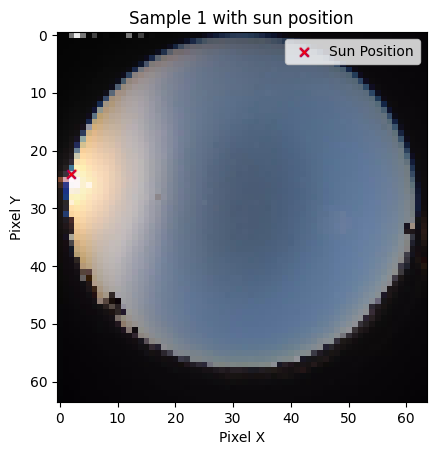}\label{fig:sub3}}&
{\includegraphics[width=0.27\linewidth]{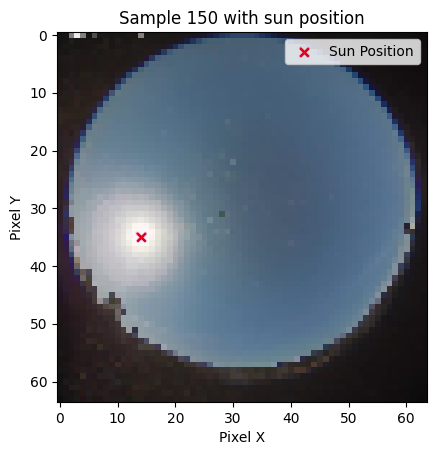}\label{fig:sub4}}
\end{tabular}}
 \caption{\textbf{Analytical sun position estimation.} Four sky input images in different conditions where our sun position estimation is displayed by means of a \textcolor{red}{$\times$} red mark. Best viewed in color.}
\label{fig:sunpos}
\end{figure}

A vertical offset correction is also applied to account for camera tilt such that:
\begin{equation}
\delta_y = y_0 + m (\epsilon - 45^\circ),
\end{equation}
with $y_0 = -3.0$ and $m = 0.03$.
The final $(x, y)$ coordinates in the image plane (origin at upper-left, $y$-axis downward) are then inferred as:
\begin{align}
x &= C + r \cdot \sin(\theta),\\
y &= C + r \cdot \cos(\theta) + \delta_y ,
\end{align}
where $C$ is the image center (e.g., 32 pixels for 64$\times$64 images). In spite of not having a ground truth for evaluation, we provide some estimations of the resulting sun position in different scenarios in Fig.~\ref{fig:sunpos}. As it can be seen, the results show some alignment trade-offs between center and corner positions, despite the empirical corrections applied throughout the transformation process. However, at least visually, the estimation seems correct and accurate to be included in our physical-aware PV power estimation model. In fact, this solution seems to be more robust than that obtained in~\cite{Berresheim_Agudo_2024}, where an image processing approach was used for this task, especially when the sky is very cloudy.

\section{Nowcast and Forecast PV power estimation}

In this paper, and following~\cite{nie2023skippd,Berresheim_Agudo_2024}, we present two neural models for nowcast and forecast PV power estimation. In the first case, our method retrieves an instantaneous power estimation from a single sky image. In addition to that, as this is a short-term prediction where high-accuracy information could be available, we may exploit real-time observations such as those coming from radars, satellites or weather stations; for rapidly evolving events like thunderstorms, heavy precipitation, or fog just to name a few. Particularly, we propose to exploit meteorological data to improve our neural model. Moreover, in this context the spatial scale is local (microscale and gamma mesoscale, few kilometers of resolution). Our model $f_N$ can be represented as:
   \begin{equation}
    f_N : (I_i,M_i) \rightarrow P_i ,
    \end{equation}
where $I_i$ represents the current image, $M_i$ the meteorological information and $P_i$ the PV power estimation at the $i$-th time. 

In the second case, the forecasting model, the PV power production of the last 16 minutes together with the corresponding sky images are considered for a 15-minute ahead prediction. Our model $f_F$ can be written as: 
\begin{equation}
f_F : (I, M, P)_{t_i - H:\delta:t_i} \rightarrow P_{t_i + T},
\end{equation}
where $H = 15$ minutes is the length of historical terms, $\delta = 1$ min the interval between historical terms, a 2 minute frequency sampling for the forecasting case, and $T = 15$ minutes represents the prediction horizon. Then, the main difference between both models is the use of information: nowcasting uses only instantaneous inputs and forecasting incorporates historical context sequences. In both cases the output power estimation is given in kilowatts (kW).

\begin{figure}[t!]
    \centering
    \resizebox{8.8cm}{!} { 
    \begin{tabular}{@{}c|c@{}}
    \includegraphics[width=1.0\linewidth]{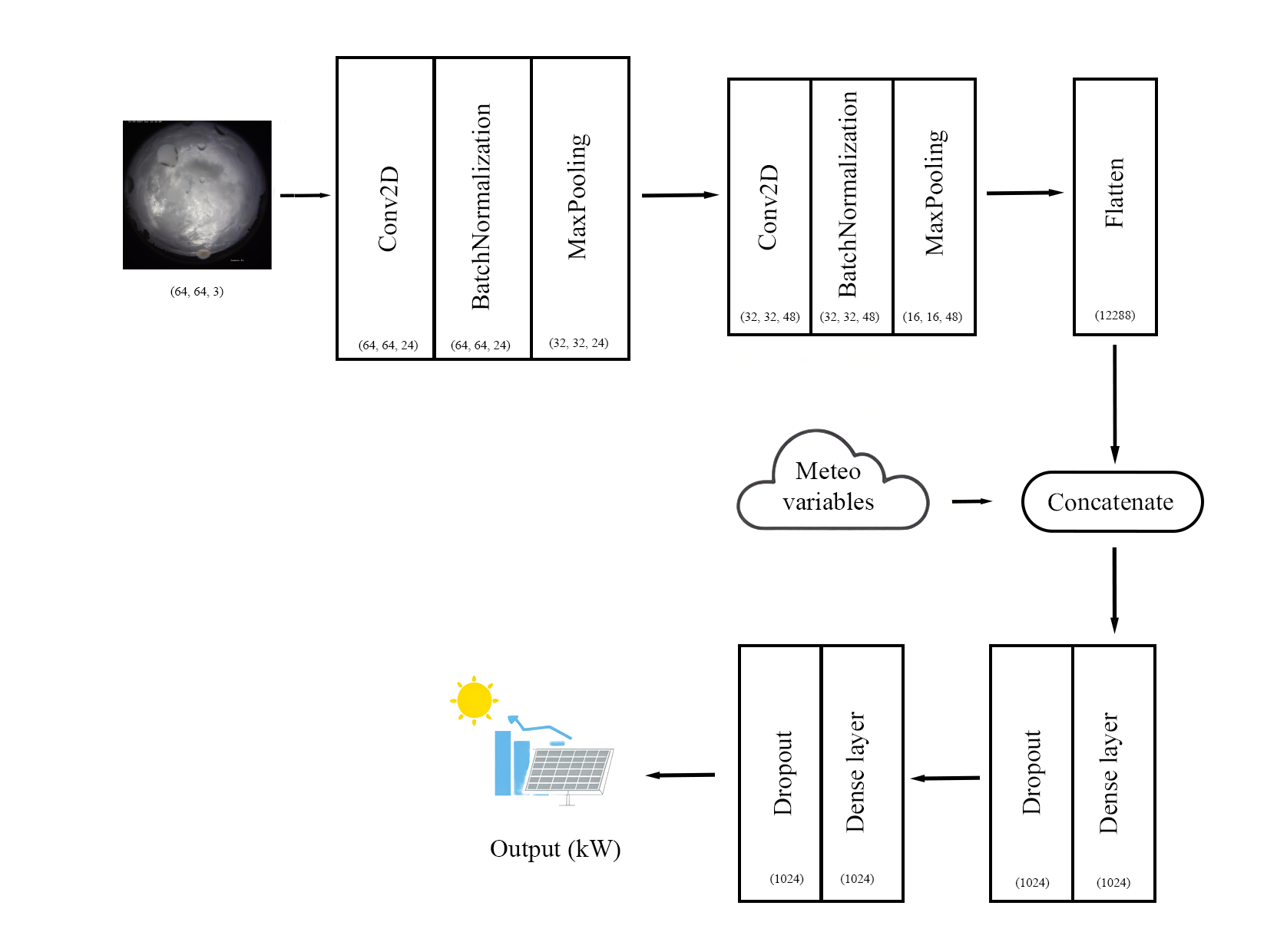}&
    \includegraphics[width=1.0\linewidth]{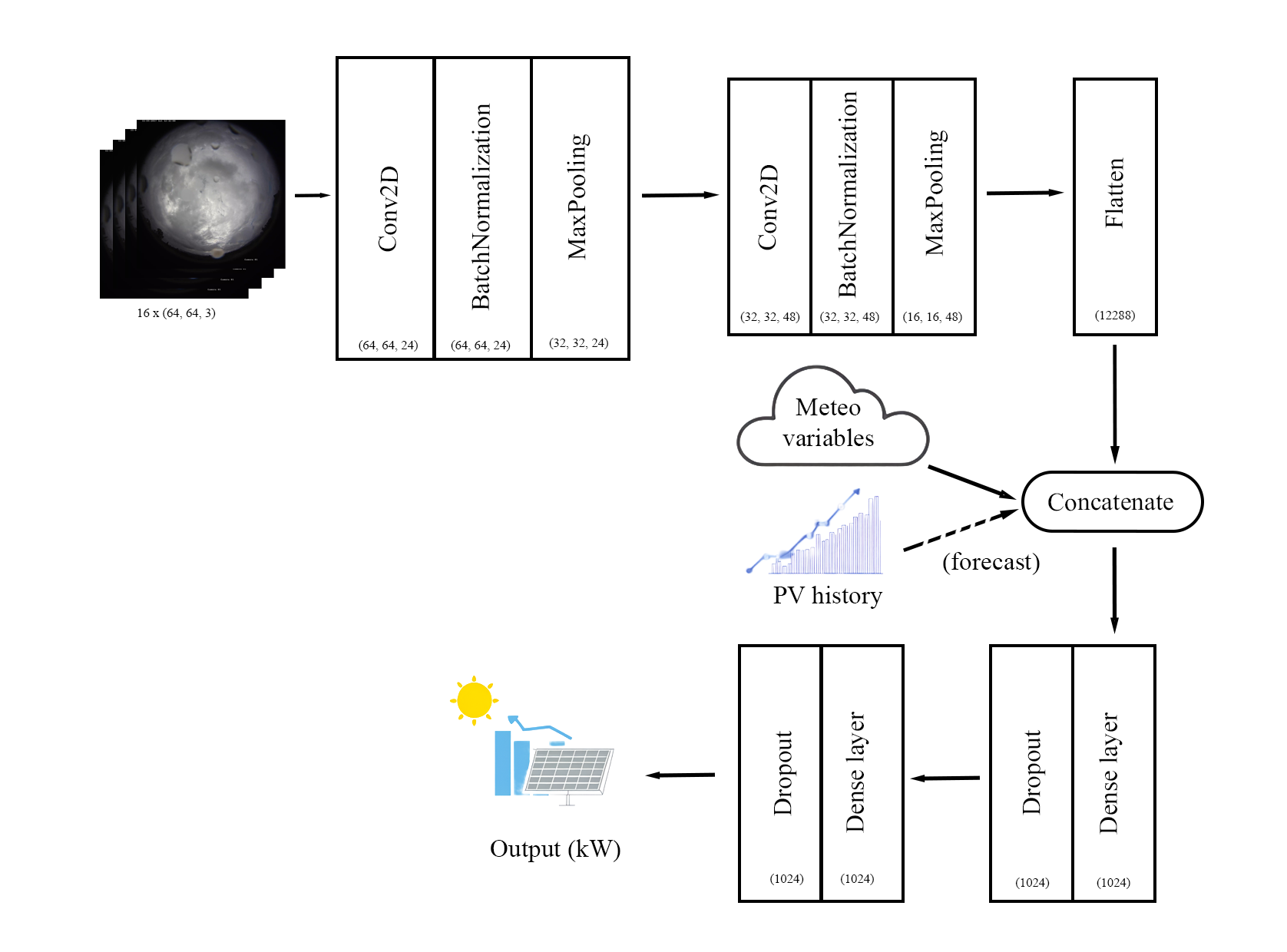}
    \end{tabular}}
    \caption{\textbf{Architectures of the neural model for both nowcast (left) and forecast (right) tasks.} The diagram shows the different blocks, the input variables fed into the model, and the output shapes at each stage.}
    \label{fig:models_representation}
\end{figure}

Fig.~\ref{fig:models_representation} shows the overall structure of the nowcast and forecast models, where the differences in the input data that feed into each architecture can be seen. Without loss of generality, both models are quite similar in terms of the main architecture we exploit. In particular, two blocks with convolutional layers with 3$\times$3 filters with a stride of 1 and same-value padding, having 24 and 48 filters, respectively; as well as batch normalization and a rectified linear unit function for activation are considered. The last part of the block being a 2$\times$2 max-pooling layer with a stride of 2. After that, flattening the tensors the data are sent to the third and last block, which consists of two fully connected layers with 1,024 neurons each and a dropout of 40\% to enhance stability of the model. Before the FC layers, this is where the difference between the two models lies, as it is here that the historical PV data are considered, along with the meteorological variables we propose to use, i.e., all variables are concatenated with flattened image features, introducing physical information directly into the latent feature space. To the best of our knowledge, this is the first time this type of information has been used in order to sort out this problem.

In particular, three sets of variables are considered to constrain both nowcast and forecast models:
\begin{itemize}
    \item \textbf{Single meteorological variable.} 4-element vector representing values from NW, NE, SW, SE grid points.
    \item \textbf{Multiple meteorological variables.} Concatenated vector of different combinations of meteorological variables, see them in section~\ref{variables_meteo} (length = 4 $\times$ number of variables).
    \item \textbf{Sun position.} Normalized coordinates \texttt{(sun\_x, sun\_y)}, with shape 1D and two values.
\end{itemize}

Both models are trained using Adam optimizer~\cite{Kingma_adam_solver_2014} with a learning rate of $3\times10^{-6}$. In training, a mean squared error loss is minimized between the ground truth and the prediction.

\section{Experimental Results}

In this section, we present our experimental evaluation for both nowcast and forecast models as well as a comparison with state-of-the-art approaches~\cite{nie2023skippd,Berresheim_Agudo_2024}. Following literature, model performance is evaluated using the Root Mean Squared Error (RMSE) and Mean Absolute Error (MAE) as:
\begin{align}
    \text{RMSE} &= \sqrt{\frac{1}{N} \sum_{n=1}^{N} (y_n - \hat{y}_n)^2},\\
    \text{MAE} &= \frac{1}{N} \sum_{n=1}^{N} |y_n - \hat{y}_n|,
\end{align}
where $y_n$ and $\hat{y}_n$ denote the $n$-th ground truth value and the prediction, respectively; and $N$ the number of samples. 

\begin{table*}[t!]
\centering
\caption{\textbf{Nowcast and Forecast PV power predictions in sunny, cloudy and overall scenarios in terms of RMSE and MAE.} Nowcast (top) and forecast (bottom) results as a function of several sources of information. The table also reports the results provided by SUNSET~\cite{nie2023skippd} and RSUNSET~\cite{Berresheim_Agudo_2024} models. $^{*}$ means RSUNSET + sun position + SAMPI results, the best solution in~\cite{Berresheim_Agudo_2024}. In \text{bold} are marked the best models.}
\resizebox{11.8cm}{!} {
\begin{tabular}{l|c|c}
\hline
\textbf{\raisebox{0.6ex}{Model}} &
\shortstack{\rule{0pt}{2.8ex}\textbf{RMSE}$\downarrow$ \\ (sunny / cloudy / overall)} &
\shortstack{\rule{0pt}{2.8ex}\textbf{MAE}$\downarrow$ \\ (sunny / cloudy / overall)} \\
\midrule
\multicolumn{3}{c}{\em Nowcast results}\\
\hline
SUNSET~\cite{nie2023skippd} & 0.804 / 3.335 / 2.428 & 0.657 / 2.337 / 1.499 \\
RSUNSET$^{*}$~\cite{Berresheim_Agudo_2024} & 0.790 / 3.300 / 2.400 & 0.650 / 2.300 / 1.480 \\
MSUNSET (Ours) & 0.804 / 3.335 / 2.428 & 0.657 / 2.337 / 1.499 \\
+\texttt{tcc} & 0.783 / 3.343 / 2.430 & 0.643 / 2.345 / 1.496 \\
+\texttt{i10fg}  & \textbf{0.764 / 3.305 / 2.401} & \textbf{0.636 / 2.302 / 1.471} \\
+\texttt{ssrd} & 0.768 / 3.355 / 2.436 & 0.640 / 2.360 / 1.502 \\
+\texttt{str} & 0.746 / 3.369 / 2.442 & 0.612 / 2.370 / 1.493 \\
+\texttt{tsr} & 0.828 / 3.332 / 2.430 & 0.678 / 2.327 / 1.504 \\
+\texttt{strd} & \textbf{0.732 / 3.316 / 2.404} & \textbf{0.600 / 2.319 / 1.461} \\
+\texttt{sp} & 0.847 / 3.333 / 2.434 & 0.708 / 2.336 / 1.524 \\
+\texttt{fdir} & 0.821 / 3.358 / 2.447 & 0.681 / 2.370 / 1.527 \\
+\texttt{Wind100} & 0.798 / 3.328 / 2.423 & 0.673 / 2.320 / 1.498 \\
+\textrm{sun position} & 0.791 / 3.372 / 2.452 & 0.645 / 2.374 / 1.512 \\
+\texttt{i10fg}+\texttt{Wind100}+\texttt{strd} & 0.864 / 3.283 / 2.403 & 0.756 / 2.266 / 1.513 \\
+\texttt{i10fg}+\texttt{Wind100} & \textbf{0.782 / 3.314 / 2.410} & \textbf{0.660 / 2.285 / 1.474} \\
+\texttt{i10fg}+\texttt{strd}+\texttt{Wind100}+\texttt{tcc} & 0.820 / 3.297 / 2.405 & 0.699 / 2.273 / 1.488 \\
+\texttt{i10fg}+\texttt{strd}+\texttt{str}+\texttt{Wind100} & 0.818 / 3.300 / 2.406 & 0.703 / 2.281 / 1.493 \\
+\texttt{i10fg}+\texttt{Wind100}+\textrm{sun position} & \textbf{0.799} / \textbf{3.297} / \textbf{2.401} & \textbf{0.689} / \textbf{2.276} / \textbf{1.484} \\
+\texttt{i10fg}+\texttt{Wind100}+\textrm{sun position}+\texttt{strd} & 0.826 / 3.302 / 2.409 & 0.707 / 2.274 / 1.493 \\
\hline
\multicolumn{3}{c}{\em Forecast results}\\
\hline
SUNSET~\cite{nie2023skippd} & 0.610 / 4.270 / 3.030 & 0.500 / 2.950 / 1.710 \\
RSUNSET$^{*}$~\cite{Berresheim_Agudo_2024} & 8.620 / 7.610 / 8.310 & 7.550 / 5.750 / 6.650 \\
MSUNSET (Ours) & 2.461 / 2.539 / 2.500 & 2.109 / 1.938 / 2.023 \\
+\texttt{i10fg} & 2.589 / 2.615 / 2.603 & 2.197 / 2.037 / 2.117 \\
+\texttt{strd} & \textbf{2.331 / 2.489 / 2.412} & \textbf{1.924 / 1.882 / 1.903} \\
+\texttt{i10fg}+\texttt{Wind100} & 2.831 / 2.669 / 2.751 & 2.402 / 2.056 / 2.229 \\
+\texttt{i10fg}+\texttt{Wind100}+\textrm{sun position} & \textbf{2.570 / 2.494 / 2.532} & \textbf{2.269 / 1.934 / 2.101} \\
+\texttt{i10fg}+\texttt{Wind100}+\texttt{strd} & 3.056 / 2.762 / 2.912 & 2.606 / 2.088 / 2.345 \\
\bottomrule
\end{tabular}}
\label{tab:results}
\end{table*}

Table~\ref{tab:results} reports our experimental results for both nowcast and forecast PV power models, including the state-of-the-art approaches SUNSET~\cite{nie2023skippd}
and RSUNSET~\cite{Berresheim_Agudo_2024} as well as our own implementation of the SUNSET model denoted as MSUNSET. Using MSUNSET as a baseline where just the sky images are considered, different one-by-one meteorological variables (see section~\ref{variables_meteo}) are exploited to finally be combined and concatenated with the sun position we propose. In general, it is worth pointing out the difference in performance between sunny (0.73/2.33kW) and cloudy (3.32/2.49kW) days for nowcast/forecast tasks, respectively, being the cloudy scenario more challenging. Next, we analyze the results in depth. 

\noindent \textbf{Nowcast results.} The variable \texttt{tsr} (top net solar radiation) improved predictions on cloudy days, contrary to intuitive expectations. This suggests that atmospheric radiation attenuation may serve as an indicator of cloud presence and density, and model would learn the interception of solar radiation before reaching the surface. Then, regarding the rest of the radiative variables (\texttt{strd}, \texttt{ssrd}, and \texttt{str}), performed better on sunny days, likely due to simplified radiative states under clear conditions.

Surface pressure (\texttt{sp}) showed limited improvement on cloudy days despite its theoretical relationship with wind patterns. In fact, \texttt{i10fg + wind100} provided a consistent improvement in prediction. In particular, the combination \texttt{i10fg + wind100 + strd} significantly worsened the results despite the improvements in individual cases, indicating the difficulty of the model in capturing complex variable interactions. However, \texttt{i10fg + wind100 + sun position} achieves one of the most accurate solutions, even outperforming competing methods~\cite{nie2023skippd,Berresheim_Agudo_2024}, suggesting that the context provided by the solar position along with other effective predictor variables increases its effectiveness.

\noindent \textbf{Forecast results.} In forecasting, we were not able to exactly reproduce the results of the forecast models reported in~\cite{nie2023skippd,Berresheim_Agudo_2024}, despite we deduced the dataset ourselves by following the procedure in the original papers. As it can be seen, our baseline provides in terms of RMSE better results in cloudy days but worse in sunny ones, probably due to the final partitioning of the data. The model's specific improvement on cloudy days, despite clear days being inherently more predictable, suggests it has effectively learned to handle the complexity and variability characteristic of cloudy scenarios. Then, the improvement in RMSE but not MAE indicates that the model has successfully reduced extreme errors, as RMSE penalizes large errors while MAE treats all errors equally.

This time, we only consider the best combinations in nowcasting. On the one hand, when our model is considering the variables (\texttt{i10fg}, \texttt{wind100}) and \texttt{sun position}, the solution is good on average, but it is not capable of surpassing the base model, suggesting that MSUNSET adapts better to cloudy conditions without variable complexity. Contrary to expectations, combining solar position with wind variables enhances forecasting in cloudy rather than sunny conditions, as opposed to the nowcast model. This unexpected result suggests both nowcast and forecast models effectively learn meaningful patterns from wind-sun position interactions. On the other hand, the best solution on average is achieved by exploiting the variable \texttt{strd}, providing a greatest improvement in cloudy days, contrasting with the performance in the nowcast model, where it excels in sunny conditions. This difference suggests that forecast models leverages historical information that allowed to better use thermal patterns associated with cloud cover, while on clear days it may be less relevant or could even interfere with the model’s pattern learning. In contrast, nowcast model, despite having one sky image available, benefit from \texttt{strd} as a stable clear-sky indicator.

\begin{figure*}[t!]
\centering
\resizebox{11.8cm}{!} { 
\begin{tabular}{@{}cc@{}}
\hspace{-0.2cm}
        \includegraphics[width=\linewidth]{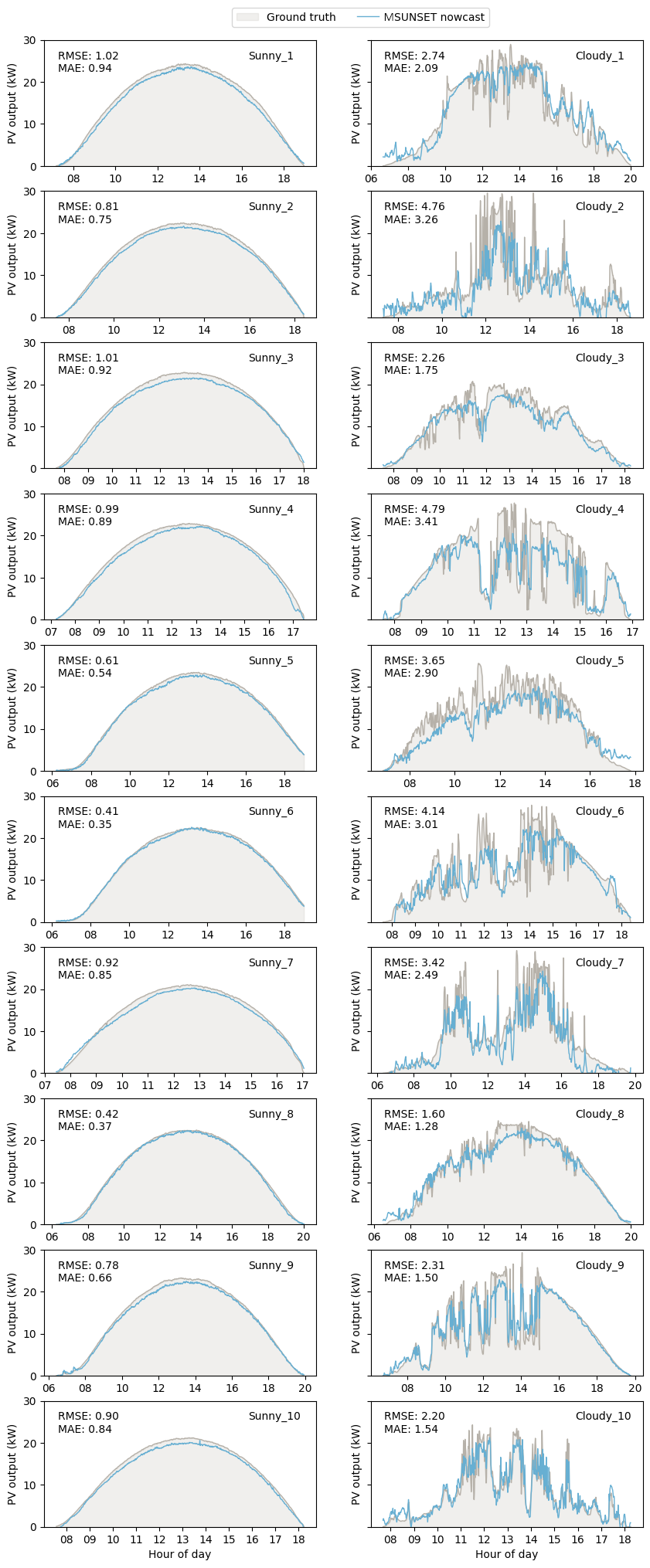}
\hspace{-0.55cm}
        \includegraphics[width=\linewidth]{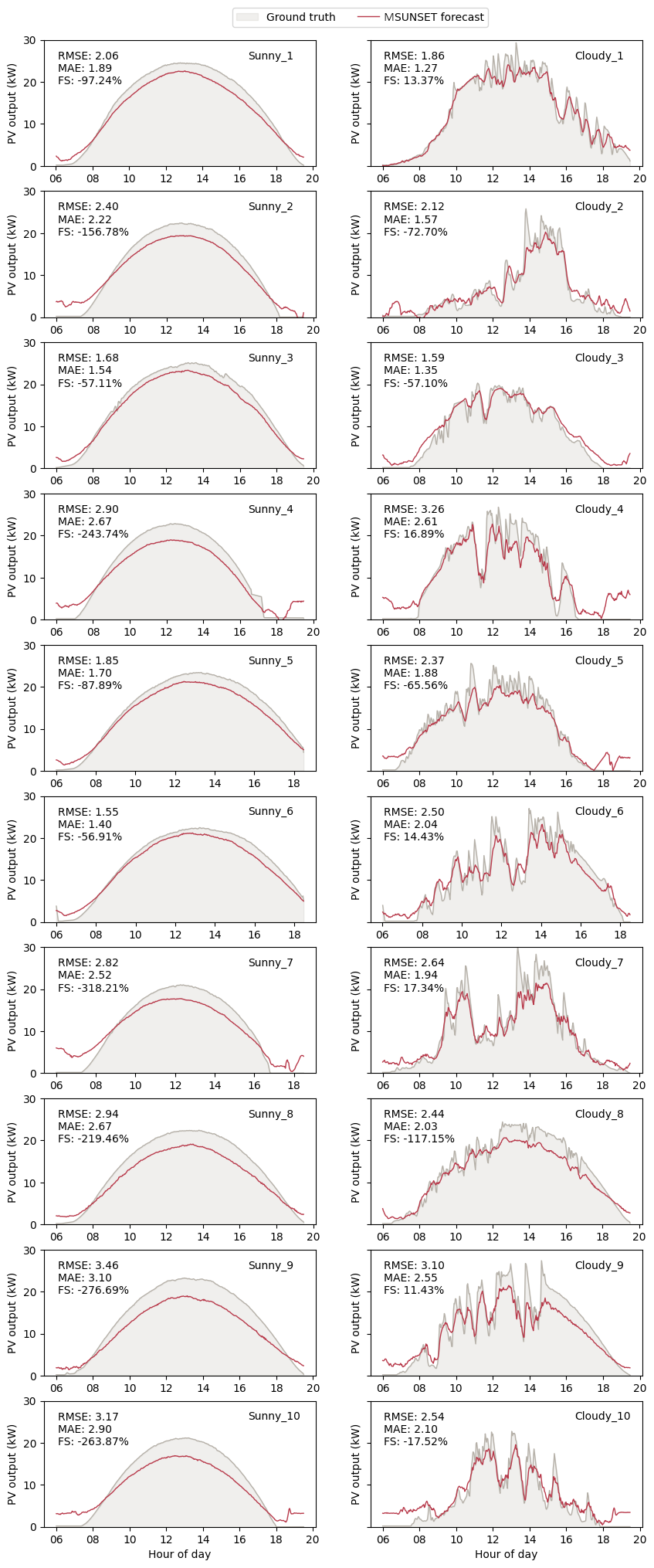}
 \end{tabular}}
 \vspace{-0.2cm}
    \caption{\textbf{Qualitative comparison on nowcast and forecast estimation for sunny and cloudy days.} In both cases, we report our estimation MSUNSET + \texttt{i10fg} + \texttt{wind100} + \texttt{sun position}. Grey area corresponds to the ground truth; blue and red lines represent the result of our model prediction for nowcast and forecast, respectively.}
    \label{fig:graphs}
 \vspace{-0.6cm}
\end{figure*}

On balance, it is concluded that the use of \texttt{strd} as well as the wind parameters (\texttt{i10fg}, \texttt{wind100}) and solar position (\texttt{sun position}) provides the most significant improvements in both nowcast and forecast tasks. A qualitative evaluation in ten test sunny and cloudy days for that combination is displayed in Fig.~\ref{fig:graphs}. As it can be seen, our algorithm provides a competitive solution in both cases, while helping to obtain a better interpretability of the solution.

\section{Conclusion}

In this paper we demonstrate the effectiveness of a hybrid multimodal approach for PV power forecasting that integrates sky images, historical power generation data, and meteorological variables through deep neural models. The results confirm that incorporating specific meteorological parameters, particularly surface long-wave (thermal) radiation downwards, wind, and solar position, enhances both short- and long-term prediction accuracy, with marked improvements specially under cloudy conditions. These findings underline the value of combining diverse data modalities to mitigate the inherent variability of solar energy production and to strengthen the robustness and interpretability of predictive models. Then, this approach contributes to more reliable grid operation and efficient management of renewable energy resources, paving the way for future developments in adaptive and data-driven solar forecasting systems.

\bibliographystyle{IEEEtran}
\bibliography{references}

\end{document}